\documentclass{article}

 \usepackage[preprint]{neurips_2026}

\usepackage[utf8]{inputenc} % allow utf-8 input
\usepackage[T1]{fontenc}    % use 8-bit T1 fonts
\usepackage{hyperref}       % hyperlinks
\usepackage{url}            % simple URL typesetting
\usepackage{booktabs}       % professional-quality tables
\usepackage{amsfonts}       % blackboard math symbols
\usepackage{nicefrac}       % compact symbols for 1/2, etc.
\usepackage{microtype}      % microtypography
\usepackage{xcolor}         % colors

\usepackage{amsthm,amsmath}
\usepackage{mathtools}

\usepackage{relsize}
\usepackage{array,booktabs,multirow}
\usepackage{xcolor}
\usepackage{colortbl} % For row colors
\usepackage{wrapfig} 
\usepackage{subcaption}
\usepackage[accsupp]{axessibility}
\definecolor{eceLow}{HTML}{76B7B2}
\definecolor{eceMid}{HTML}{BAB0AC}
\definecolor{eceHigh}{HTML}{E15759}
\definecolor{eceLine}{HTML}{5B2A86}
\definecolor{regColor}{HTML}{A9D18E}
\definecolor{zsColor}{HTML}{26547C}
\definecolor{oursColor}{HTML}{E86AA3}
\usepackage{enumitem}
\title{Rethinking the Test-Time Prompt Tuning Objective from the Perspective of Calibration}
\author{%
  Jungwon Choi$^{1,*}$ \quad
  Hyeonseo Jang$^{2,*}$ \quad
  Kibok Lee$^{2}$ \quad
  Eunwoo Kim$^{1}$ \\[1ex]
  $^{1}$Department of Computer Science and Engineering, Chung-Ang University \\
  $^{2}$Department of Statistics and Data Science, Yonsei University \\[0.5ex]
  $^{*}$These authors contributed equally.
}

\begin{document}

\maketitle

\begin{abstract}

Test-time prompt tuning (TPT) has emerged as a powerful paradigm, refining prompts for each test sample via entropy minimization (EM) over multiple augmented views. 
However, we identify a limitation in the standard EM-based adaptation: it inherently drives the model toward overconfident predictions disregarding sample-specific uncertainty, leading to significant calibration degradation.
% Furthermore, ~ 는 근본적으로 정교한 prediction contrl을 어렵게 함  (??)
Moreover, EM in TPT operates on augmented views rather than the original view used for inference, making it difficult to effectively control prediction confidence.
%This issue is exacerbated by a structural mismatch, where the EM loss is applied to the augmented views while the final inference relies on the original (i.e., canonical) view.
Unfortunately, existing approaches retain the EM objective and rely on additional regularization terms, leaving the fundamental issue unresolved.
%To resolve these limitations, we propose a new objective to replace the conventional EM loss.
%Our method directly aligns the original-view prediction with a target distribution derived from augmented views through a cross-entropy, while simultaneously incorporating a target-entropy subtraction term. 
%To address these limitations, we propose a new objective as an alternative to the conventional EM loss. Specifically, our method directly aligns the original-view prediction with a target distribution derived from augmented views through cross-entropy, while adversarially incorporating the entropy of the target distribution to capture sample-specific uncertainty.
To address these limitations, we propose a new objective that replaces the conventional EM loss by aligning the original-view prediction with a target distribution derived from augmented views via cross-entropy, while adversarially incorporating the entropy of the target distribution to capture sample-specific uncertainty.
Furthermore, to better construct this target distribution, we apply confidence-aware temperature scaling to each augmented-view prediction according to its confidence, sharpening confident predictions while softening uncertain ones.
% This formulation allows the model to selectively increase confidence when augmented views yield consistent predictions, while maintaining low confidence when they are inconsistent. 
%This formulation enables the model to increase confidence when the target distribution is confident, while preserving low confidence when the target distribution exhibits high uncertainty.
This formulation allows the model to increase confidence only when the target distribution is reliable, while preserving uncertainty when it reflects ambiguous or conflicting augmented-view predictions.
%Furthermore, we introduce Confidence-aware Adaptive Temperature Scaling (CATS), which assigns lower temperatures to confident views and higher temperatures to uncertain ones, producing sharp or softened distributions accordingly.
%This strategy ensures that the aggregated target distribution more faithfully reflects the underlying uncertainty of the augmentations.
Extensive experiments across diverse benchmarks demonstrate that our approach not only achieves state-of-the-art accuracy but also significantly improves model calibration.
\end{abstract}

\section{Introduction}
\label{sec:intro}

% Pre-trained vision-language models such as CLIP~\citep{clip} demonstrated remarkable zero-shot transferability by aligning images and texts in a shared embedding space. Building on this capability, prompt tuning~\citep{coop, cocoop, maple} emerged as an effective way to adapt these models by optimizing textual prompts instead of updating the entire model. Recently, test-time prompt tuning (TPT)~\citep{tpt} has attracted considerable attention because it enables sample-wise adaptation using only unlabeled test data, making it especially appealing for realistic deployment scenarios where labeled target data are unavailable. In particular, TPT improves zero-shot generalization by refining prompts at inference time through an entropy minimization (EM) objective computed from multiple augmented views of a test image~\citep{difftpt, dynaprompt, c-tpt, o-tpt}. This simple adaptation strategy has proven effective in improving prediction accuracy under distribution shift and has inspired a growing body of follow-up studies.

Pre-trained vision-language models, such as CLIP~\citep{clip}, demonstrated remarkable zero-shot transferability by aligning images and texts in a shared embedding space. 
Building on this capability, prompt tuning~\citep{coop, cocoop, maple} emerged as an effective way to adapt these models by optimizing textual prompts instead of updating the entire model. 
Recently, test-time prompt tuning (TPT)~\citep{tpt} has attracted considerable attention by enabling adaptation using only unlabeled test samples, which is achieved by refining prompts at inference time with an entropy minimization (EM) loss over prediction probabilities computed from multiple augmented views~\citep{difftpt, dynaprompt, c-tpt, o-tpt}. This simple adaptation objective has proven effective in improving downstream task performance under distribution shift and has inspired a growing body of follow-up studies~\cite{difftpt, selftpt, dynaprompt, choi2026dual, fpp}.

%%%%%%%%%%%%%%%%%%%%%%%%%%%%%%%%%%%%%%%%%%%%%%%%%%%%%%%%%%%%%%%%%%%%%%%%%%%%%%%%%%%%%%%%%%%%%%%%%%%%%%%%%%%%%%%%%%%%%%%%%%%%%%%%%%%%%%%%%%%%%%%%%%%%%%%%

% Despite its effectiveness, accuracy alone is insufficient for trustworthy deployment. In many real-world applications, including safety-critical settings, a model must not only make i) correct predictions but also assign ii) confidence scores that faithfully reflect the likelihood of correctness. This property, known as calibration, is particularly important for vision-language models that are increasingly used in decision-making pipelines.~\citep{Liu2023,Basu2025,Khandelwal2022} However, prior work has shown that TPT often degrades calibration even when it improves accuracy. A key reason is that its entropy minimization objective consistently encourages sharper prediction distributions, which can easily lead to overconfident outputs as shown in Section~\ref{sec:morestudies}. As a result, TPT may produce predictions that appear highly certain even when they are unreliable, making calibration another central challenge in test-time prompt tuning. 

Despite its effectiveness, however, the EM loss used in TPT has an inherent limitation for trustworthy deployment. 
As analyzed in Section~\ref{sec:morestudies}, it consistently encourages sharper prediction probabilities regardless of the characteristics of each test sample. 
Although this tendency can improve prediction accuracy, it may also increases the mismatch between prediction correctness and confidence scores, thereby degrading model calibration. 
Given that vision-language models are increasingly used in real-world decision-making pipelines including safety-critical settings, such degradation in calibration is particularly concerning because confidence estimates are expected to faithfully reflect the likelihood of correctness~\citep{Liu2023,Basu2025,Khandelwal2022}.

%%%%%%%%%%%%%%%%%%%%%%%%%%%%%%%%%%%%%%%%%%%%%%%%%%%%%%%%%%%%%%%%%%%%%%%%%%%%%%%%%%%%%%%%%%%%%%%%%%%%%%%%%%%%%%%%%%%%%%%%%%%%%%%%%%%%%%%%%%%%%%%%%%%%%%%%

% Recent studies have attempted to address this issue from several perspectives, including enlarging text-feature dispersion~\citep{c-tpt}, enforcing angular separation between class text features~\citep{o-tpt}, and initializing prompts in flatter regions of the loss landscape~\citep{fpp}.
% While these methods improve calibration to a meaningful extent, they share a common critical limitation: they retain the original EM loss and seek to mitigate its side effects through regularization strategies.
% Consequently, they do not resolve the fundamental mismatch between the optimization objective and calibration.
% Moreover, the standard EM loss in TPT is applied to the average prediction distribution obtained from augmented views, whereas the final prediction confidence at test time is determined by the original view.
% We empirically observe that adjusting the probabilities of augmentation views does not necessarily yield appropriate control over the probability of the original view, which is the quantity that ultimately matters for prediction confidence.
% This mismatch makes precise calibration inherently difficult and suggests that improving calibration requires revisiting the objective itself rather than merely regularizing it.

Recent studies have attempted to address this issue from several perspectives, including enlarging text-feature dispersion~\citep{c-tpt}, enforcing angular separation between class text features~\citep{o-tpt}, and initializing prompts in flatter regions of the loss landscape~\citep{fpp}.
While these methods improve calibration to a meaningful extent, they share a common critical limitation: they retain the original EM loss, the fundamental source of calibration degradation, and merely mitigate its side effects through regularization, leaving the underlying issue unresolved.
Moreover, the standard EM loss in TPT is applied to the average prediction distribution obtained from augmented views, whereas the final prediction confidence at test time is determined by the original view. 
We empirically observe that adjusting the probabilities of augmentation views does not necessarily yield appropriate control over the probability of the original view, making precise calibration inherently difficult and suggesting the need to revisit the objective itself.
%We empirically observe that adjusting the probabilities of augmentation views does not necessarily yield appropriate control over the probability of the original view.
%This mismatch makes precise calibration inherently difficult and suggests that improving calibration requires revisiting the objective itself.

%%%%%%%%%%%%%%%%%%%%%%%%%%%%%%%%%%%%%%%%%%%%%%%%%%%%%%%%%%%%%%%%%%%%%%%%%%%%%%%%%%%%%%%%%%%%%%%%%%%%%%%%%%%%%%%%%%%%%%%%%%%%%%%%%%%%%%%%%%%%%%%%%%%%%%%%

% (1) Motivated by this observation, 우리는 augmentation view의 추가 정보를 최종 prediction confidence에 직접 반영하는 동시에, 추가 정보에 내재된 uncertainty를 통해 adaptation 강도를 조절하는 obejective를 제안한다.
% (2) Specifically, 우리의 loss는 treat the probability from the original view as a source distribution and construct a target distribution from augmentation views를 하고, (i) 둘 사이의 cross entropy를 minimize 하는 term을 통해 최종 예측에 쓰이는 original view confidence를 직접 update 하고 (ii) 여기에 target distribution의 entropy를 빼서 불확실성에 따라 update의 양이 조절되게 만들었다. 
% (3) Notably, 항상 entroy를 증가시키는 EM loss와 다르게 우리의 objective는 augmentation view의 정보를 통해 맞출 수 있는 sample들에서는 confidence를 올리고 모델이 맞추기 힘든 sample들에서는 confidence를 낮춰서 
% (4) 우리의 새로운 objective는 여러 downstream task에서 performance와 calibration 모두에서 동시에 SOTA results를 달성했다.

Motivated by this observation, we propose a new uncertainty-aware objective that directly incorporates the information from augmented views into the original-view prediction confidence while using the uncertainty of this information to modulate the adaptation strength.
Specifically, we treat the prediction probabilities from the original view as a source distribution to be updated and construct a target distribution from augmented views.
Furthermore, to construct a more reliable target distribution, we apply confidence-aware temperature scaling to each augmented-view prediction, sharpening confident predictions while softening uncertain ones.
Our objective consists of a cross-entropy matching term and an adversarial target-entropy regularization term.
The former transfers class-level evidence from augmented views to the original-view confidence used for the final prediction, while the latter modulates the update strength according to the uncertainty of the augmentation-based target.
%
% Temp. scaling 얘기 짧고 간결하게.
%Furthermore, to better construct this target distribution, we apply confidence-aware temperature scaling to each augmented-view prediction according to its confidence, sharpening confident predictions while softening uncertain ones.
Notably, unlike the conventional EM loss, which consistently sharpens prediction probabilities, our objective adjusts confidence according to augmentation-derived uncertainty: our experiments show that it increases confidence when augmented views provide consistent class-level evidence, while suppressing overconfident predictions when they are ambiguous or inconsistent.
As a result, the proposed objective simultaneously improves performance and model calibration, achieving state-of-the-art results across diverse benchmarks. Our contributions are summarized as follows:
\begin{itemize}[leftmargin=*, itemsep=1pt, topsep=2pt]
    \item Through a comprehensive set of experiments and analyses, we systematically examine an inherent limitation of the conventional EM loss in TPT, revealing that its optimization objective is structurally misaligned with calibration.
    %\item We analyze the inherent limitation of the conventional EM loss in TPT, demonstrating that its consistent sharpening behavior can improve accuracy but often harms calibration.
    \item We propose a new uncertainty-aware objective that directly incorporates information  from augmented views into the original-view prediction while modulating the adaptation strength based on augmentation-derived uncertainty.
    \item We achieve state-of-the-art results in both calibration and downstream task performance, without relying on additional regularization to compensate for the side effects of EM loss.
\end{itemize}

\section{Related Work}
\label{sec:relatedwork}

\textbf{Prompt Learning.}
Vision-language models (VLMs)~\citep{clip, align} have demonstrated strong generalization ability by aligning image and text representations in a shared embedding space. 
To further improve downstream performance, prompt learning methods have been proposed to adapt prompts to specific tasks~\citep{coop, cocoop, maple}. 
%For example, CoOp~\citep{coop} formulates text prompts as learnable continuous vectors and optimizes them using labeled data. 
%Building on this, CoCoOp~\citep{cocoop} introduces instance-conditioned prompts to enhance generalization to unseen classes. 
%More recently, MaPLe~\cite{maple} extends prompt learning to both visual and textual modalities, enabling deeper cross-modal interactions.
While prior prompt learning methods rely on labeled data during training, test-time prompt tuning (TPT)~\citep{tpt} updates prompt parameters using only unlabeled test samples by leveraging multiple augmented views of a single image. 
Specifically, TPT selects confident views and minimizes the entropy of their aggregated predictions, encouraging consistent outputs across augmentations.
However, existing TPT methods~\citep{tpt, c-tpt, o-tpt} primarily rely on entropy minimization as the optimization objective, assuming that selected predictions are equally reliable. As a result, the update is dominated by augmented views without explicitly anchoring the original prediction, and the entropy minimization objective can degrade calibration by promoting overconfident outputs.

\textbf{Calibration of Deep Neural Networks.}
Calibration evaluates how well the predicted confidence of the model aligns with its empirical accuracy~\citep{Guo2017}, which is critical for applications requiring reliable uncertainty estimates, such as healthcare~\citep{Liu2023, Wang2022} and autonomous systems~\citep{Bucker2022, Khandelwal2022}. 
Existing calibration approaches can be broadly categorized into post-hoc~\citep{Guo2017, Lei2018, Platt1999} and train-time methods~\cite{Karandikar2021, Kumar2018, Yoon2023}. 
Post-hoc techniques, such as temperature scaling~\cite{Guo2017} and Platt scaling~\cite{Platt1999}, adjust a trained model using a held-out validation set to better align predicted probabilities with observed outcomes. 
However, they rely on labeled data from the target distribution, limiting their effectiveness under distribution shift or in zero-shot settings~\citep{Liu2022b}. 
Train-time methods incorporate additional objectives or regularization to encourage calibrated predictions~\citep{Karandikar2021, Kumar2018, Yoon2023}. 
While effective in supervised settings, these methods also require labeled data and cannot be directly applied to test-time adaptation, where only unlabeled test samples are available.

\textbf{Calibration of VLMs.}
While VLMs have demonstrated strong performance across diverse tasks, they often suffer from poor calibration when adapted to new domains, producing overconfident predictions. 
TPT~\citep{tpt} improves task-specific accuracy, but can further exacerbate this issue due to its entropy minimization objective. 
Recent works attempt to improve calibration through various strategies. For instance, C-TPT~\citep{c-tpt} links textual feature dispersion to calibration and proposes a loss to enlarge inter-class separation, while O-TPT~\citep{o-tpt} introduces orthogonality-based regularization to enhance angular separation. 
% Another line of work, such as FPP~\citep{fpp}, focuses on initializing prompts in flatter regions of the loss landscape. 
However, these methods still rely on entropy minimization and only partially mitigate its side effects, without addressing the root cause of calibration degradation. 
Moreover, the entropy loss is applied to aggregated predictions from augmented views, whereas the final prediction confidence is determined by the original view, leading to a fundamental mismatch between the optimization objective and calibration. 
In contrast, our approach moves beyond existing entropy minimization methods by directly aligning the original-view prediction with a confidence-aware target constructed from augmented views, enabling more explicit control over prediction confidence at inference time.
\section{Preliminaries}
\label{sec:method}

\textbf{Zero-shot Classification.} We first introduce a vision-language model~\citep{clip} that aligns images and texts in a joint embedding space via contrastive learning. For zero-shot classification, given a set of classes $\mathcal{C} = \{c_1, c_2, \dots, c_K\}$, each class name is embedded into a textual prompt $t_{c_i}$ using a template (e.g., ``a photo of a \{class\}''). The text encoder extracts the corresponding text embeddings,
$
    e_{c_i} = f_T(t_{c_i}), \text{for } i = 1, \dots, K.
$
A test image $I$ is mapped to a visual embedding $v = f_I(I)$. The alignment between the image and each class is measured using cosine similarity, $s_i = \cos(v, e_{c_i})$. These scores are converted into a probability distribution via the Softmax function,
$
    P(c_i|I) = \frac{\exp(s_i / \tau)}{\sum_{j=1}^{K} \exp(s_j / \tau)},
$
where $\tau$ is a temperature parameter that controls the sharpness of the distribution. The final prediction $\hat{c}$ and its associated confidence $\hat{P}$ are determined as
$
    \hat{c} = \arg\max_{c_i} P(c_i|I)$ and $ \hat{P} = \max_{c_i} P(c_i|I).
$
This mechanism allows the model to perform robust classification across open-vocabulary tasks without requiring additional task-specific fine-tuning.

\textbf{Test-time Prompt Tuning (TPT)}~\citep{tpt} aims to adapt prompt parameters $\theta$ at inference time to enhance model robustness under distribution shifts without labeled data. Specifically, TPT optimizes $\theta$ by minimizing the prediction entropy across multiple augmented views of a single test image $I$. To ensure high-quality guidance for prompt, it filters these augmented views based on a confidence threshold and computes an aggregated probability distribution $\tilde{p}_{\theta}$ from the selected low-entropy views. The prompt is then updated by minimizing the entropy of this aggregated prediction:
\begin{equation}
    \mathcal{L}_{TPT} = - \sum_{i=1}^{K} \tilde{p}_{\theta}(y_i|I) \log \tilde{p}_{\theta}(y_i|I),
\end{equation}
where $K$ denotes the total number of classes. This objective encourages the model to produce consistent and confident predictions across augmentations.

% \textbf{Calibration Performance and Metric.}
% Following previous studies, we assess model calibration using two key metrics: Expected Calibration Error (ECE) and Static Calibration Error (SCE).

% The ECE quantifies the alignment between predicted confidence and empirical accuracy by partitioning the total $M$ predictions into $B$ equally-spaced bins. It is defined as the weighted average of the absolute difference between accuracy and confidence across all bins:
% \begin{equation}
%     \text{ECE} = \sum_{b=1}^{B} \frac{|A_b|}{M} \left| \text{acc}(A_b) - \text{conf}(A_b) \right|,
% \end{equation}
% where $A_b$ represents the set of samples in the $b$-th bin, while $\text{acc}(A_b)$ and $\text{conf}(A_b)$ denote the average accuracy and predicted confidence within that bin, respectively. 

% To provide a more comprehensive evaluation in multi-class scenarios, we also employ the SCE. Unlike ECE, which only considers the maximum confidence score, SCE calculates the calibration error across all classes $K$ and all bins $B$:
% \begin{equation}
%     \text{SCE} = \frac{1}{K} \sum_{k=1}^{K} \sum_{b=1}^{B} \frac{|A_{k,b}|}{M} \left| \text{acc}(A_{k,b}) - \text{conf}(A_{k,b}) \right|,
% \end{equation}
% where $A_{k,b}$ denotes the subset of samples belonging to class $k$ whose predicted confidence for that class falls within the $b$-th bin.

\begin{figure*}[t] % 위치 옵션: t=top, b=bottom, h=here, H=exactly here  
    \centering
    \includegraphics[width=\linewidth,
    trim=0pt 140pt 0pt 0pt,  
    clip=true]{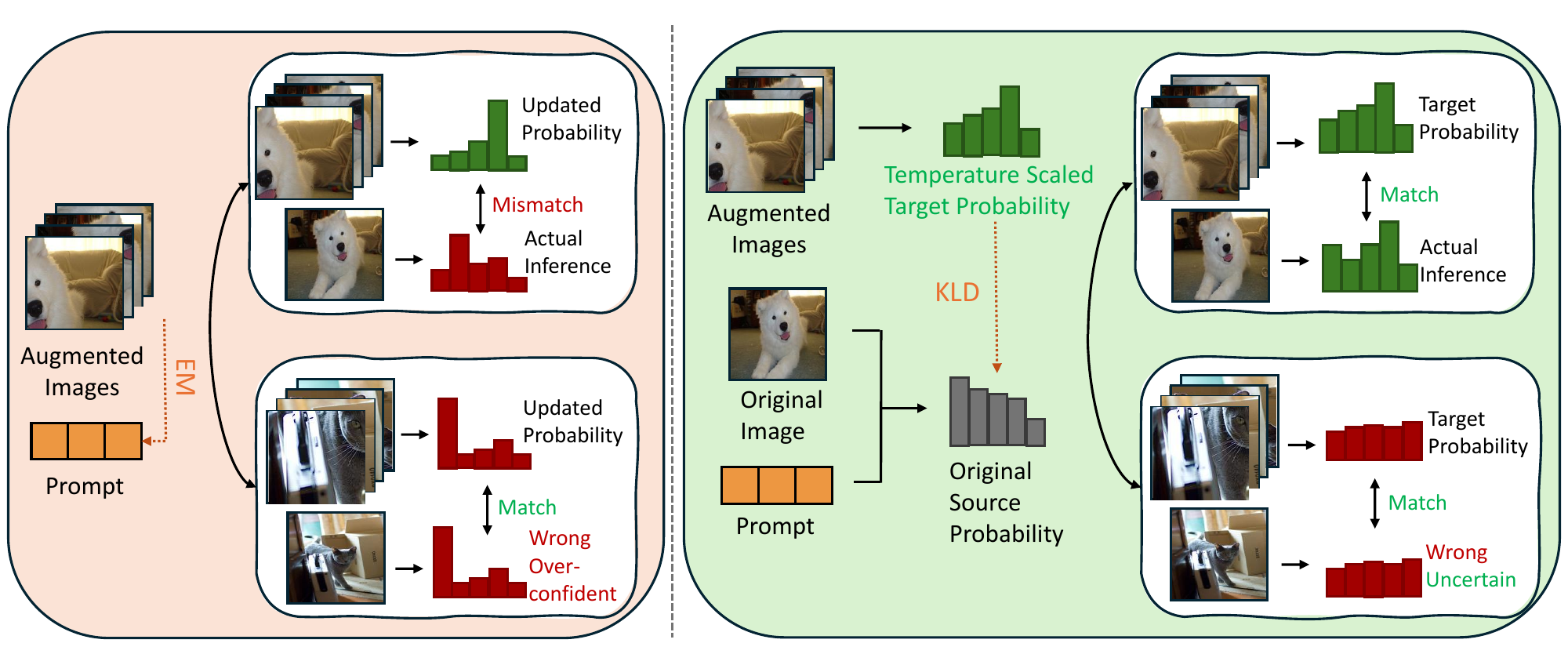}
    \vspace{-10pt}
%     \caption{The left illustrates inference with prompts updated via EM, where predictions derived from augmented views can be misaligned with the original image, often leading to mismatched (upper left) or over-confident incorrect predictions (lower left).
% The right shows our method, which directly updates prompts using Confidence-aware Adaptive Temperature Scaling (CATS) applied to augmented views target. This results in better alignment between the target and original-view predictions (upper right), while maintaining uncertainty for ambiguous or incorrect samples. (lower right)}
\caption{Comparison between entropy minimization (EM) (left) and our objective (right). 
Entropy minimization on augmented views can lead to misalignment between augmented and original-view predictions (upper left) and produce over-confident incorrect predictions even when they are aligned (lower left).
In contrast, our proposed method constructs a confidence-aware target distribution from augmented views and directly aligns it with the original-view prediction, improving consistency between augmentation-derived targets and inference-time predictions (upper right) while producing smoother and more uncertain predictions for ambiguous samples (lower right).}
% \caption{Comparison between entropy minimization (EM) (left) and our objective (right). The upper row compares prediction consistency between augmented views and the original view used for inference: EM often produces misaligned predictions, whereas our method achieves better alignment through confidence-aware target matching. 
% The lower row compares confidence behavior on ambiguous or incorrect samples: EM generates over-confident incorrect predictions, while our method preserves uncertainty by constructing smoother target distributions from inconsistent or ambiguous augmented views, leading to improved calibration.}
    \label{fig:method}

\end{figure*}

\section{Analysis of Uncertainty-Aware Adaptation Objective}
%Rethinking the TPT Objective
%ㅎㅇ
% 목표 1: CE, -EM 각각의 필요성을 설명한다.
% 목표 2: Ours = CE - EM 으로 자연스럽게 넘어간다.
% 목표 3: Gradient 측면에서 이 효과를 수학적으로 분석한다. 이 때 일반 D_KL과 다른 점을 강조한다.

For each test sample, we use \(p\) to denote the prediction probability from the original view, which are used to obtain the final prediction and confidence at inference time.
We further define \(p_{\mathrm{aug}}\) as the target probability obtained by averaging the prediction probabilities over multiple augmented views.
Standard TPT minimizes the entropy of \(p_{\mathrm{aug}}\), thereby leveraging information aggregated across augmented views.
However, since TPT does not explicitly optimize \(p\), the prediction probabilities used for inference can still be misaligned with \(p_{\mathrm{aug}}\) after adaptation. To address this, we first consider the cross-entropy matching term:
\begin{equation}
\mathcal{L}_{\mathrm{match}}
=
H(p_{\mathrm{aug}},p)
=
-\sum_{c=1}^{C} p_{\mathrm{aug}}^{(c)} \log p^{(c)} .
\end{equation}
Minimizing this term encourages \(p\) to assign high probability to the classes tha   t receive high probability under \(p_{\mathrm{aug}}\), so that the class-level information estimated from multiple augmented views is directly reflected in the probabilities.

Beyond class-level guidance, \(p_{\mathrm{aug}}\) can also be viewed as capturing sample-dependent uncertainty.
If the augmented views give consistent predictions, the averaged distribution become sharp; if they are ambiguous or inconsistent, the averaged distribution remain soft. 
This uncertainty should be preserved for reliable confidence estimation, since an ambiguous target should not make \(p\) overconfident. We quantify it by
\begin{equation}
H(p_{\mathrm{aug}})
=
-\sum_k p_{\mathrm{aug}}^{(k)}
\log p_{\mathrm{aug}}^{(k)} .
\end{equation}
Unlike the conventional EM objective, we do not use \(H(p_{\mathrm{aug}})\) to make the target probabilities sharper.
Instead, we use it to adjust the matching cost according to the uncertainty of \(p_{\mathrm{aug}}\).
When \(p_{\mathrm{aug}}\) is sharp, its entropy is small, so the cross-entropy term remains a strong matching signal. 
When \(p_{\mathrm{aug}}\) is soft, its entropy is large, so the matching cost is reduced, preventing an uncertain target from forcing \(p\) to become overconfident.
This leads to the objective 
\begin{equation}
\mathcal{L}
=
H(p_{\mathrm{aug}},p)
-
H(p_{\mathrm{aug}}).
\end{equation}
Here, subtracting the target entropy removes the uncertainty inherent in \(p_{\mathrm{aug}}\) from the matching cost. 
As a result, the objective encourages \(p\) to follow the class preference of \(p_{\mathrm{aug}}\) when the augmented views are confident, while retaining softer predictions when the augmented views are uncertain.

The objective derived above has a compact interpretation. Expanding the two terms gives
\begin{equation}
\mathcal{L}
=
-\sum_k p_{\mathrm{aug}}^{(k)} \log p^{(k)}
+
\sum_k p_{\mathrm{aug}}^{(k)} \log p_{\mathrm{aug}}^{(k)}
=
\sum_k p_{\mathrm{aug}}^{(k)}
\log
\frac{p_{\mathrm{aug}}^{(k)}}{p^{(k)}} .
\end{equation}
Thus, the proposed objective can be written as
\begin{equation}
\mathcal{L}_{\mathrm{KLD}}
\triangleq
D_{\mathrm{KL}}(p_{\mathrm{aug}} \,\|\, p).
\end{equation}
In our setting, \(p_{\mathrm{aug}}\) is not a fixed target but remains differentiable and allows gradients to update the model.
Therefore, the entropy term \(H(p_{\mathrm{aug}})\) is not constant and contributes to the update together with the cross-entropy term. 
The negative entropy term counteracts the tendency of the cross-entropy term to produce overconfident predictions, allowing uncertain augmented targets to remain soft during adaptation.

Compared with standard TPT~\citep{tpt}, which minimizes the entropy of the averaged probabilities and can sharpen them even for ambiguous samples, our objective uses the same augmented-view information to update the probabilities used for inference while retaining target-side uncertainty. It also differs from calibration-oriented methods~\cite{o-tpt, c-tpt, fpp} that keep the standard EM objective and add a separate regularizer, since since it preserves uncertainty within the objective itself, rather than through an additional regularizer. The qualitative difference between entropy minimization and our objective is illustrated in Fig.~\ref{fig:method}

\section{Confidence-aware Adaptive Temperature Scaling}

% To better reflect the underlying uncertainty of the augmentations, we propose a Confidence-aware Adaptive Temperature Scaling strategy to construct reliable supervision signals.
% To construct the uncertainty-aware target distribution introduced in the previous section, we propose a Confidence-aware Adaptive Temperature Scaling strategy.
The above analysis suggests that confident and consistent augmented views tend to produce sharper aggregated targets, whereas uncertain or inconsistent views yield softer ones. However, individual augmented views may exhibit substantially different confidence levels. 
% Directly averaging their probability distributions treats all views equally, regardless of their reliability, allowing uncertain views to disproportionately influence the aggregated target. 
Directly averaging their probability distributions treats all views equally, without considering differences in confidence across views. To address this issue, we propose a Confidence-aware Adaptive Temperature Scaling strategy, which adjusts the sharpness of each augmented-view distribution according to its confidence before aggregation.
% augmented views는 모두 equally reliable하지 않다. 기존 방식들은 단순히 모든 뷰들의 평균 probability를 활용하지만 우리는 analysis에서 언급했듯이 objective에서 $p_{aug}$의 adversarial target entropy term을 활용하므로 aug view들에 unerly 되어있는 uncertainty를 더 잘 활용하기 위해 각 view들의 distribution을 조절한다.
Given a set of augmented views of a test image, the model first produces a probability distribution for each view, $p_i = \mathrm{softmax}(z_i),$
where $z_i$ denotes the logits of the $i$-th view.
We estimate the confidence of each view using the maximum predicted probability,
$
\alpha_i = \max_k p_i^{(k)}.
$
Based on this confidence, we assign a view-specific temperature using linear interpolation,
$
T_i = T_{\max} - (T_{\max} - T_{\min}) \cdot \alpha_i,
$
where $T_{\min}$ and $T_{\max}$ denote the minimum and maximum temperature values. 
This ensures that confident predictions produce sharper distributions, whereas uncertain predictions yield smoother distributions.
We then generate calibrated soft labels for each view,
$
\tilde{p}_i = \mathrm{softmax}\left(\frac{z_i}{T_i}\right).
$
The final target distribution is obtained by averaging across all views:
$
p_{\mathrm{aug}} = \frac{1}{N} \sum_{i=1}^{N} \tilde{p}_i,
$
where $N$ is the number of augmented views.
Finally, we minimize the KL divergence between the prediction of the original image and the aggregated target
$
\mathcal{L}_{\mathrm{KLD}}
=
D_{\mathrm{KL}}(p_{\mathrm{aug}} \,\|\, p).
$

\section{Experiments} 
\label{sec:Experiments}

\subsection{Experimental Setting}\noindent

\noindent\textbf{Implementation Details.}
We use the CLIP-ViT-B/16 architecture as the backbone. 
% For test-time adaptation, we follow the TPT~\cite{tpt} setting adopted in the baseline works~\cite{c-tpt, o-tpt} without modification, and follow O-TPT~\cite{o-tpt} for other unspecified details.
For fair comparison, we adopt the same test-time adaptation setting as TPT~\cite{tpt}, following the protocols used in prior works~\citep{c-tpt, o-tpt} without modification.
Following the baseline works~\citep{c-tpt, o-tpt}, we evaluate calibration using expected calibration error (ECE)~\citep{Naeini2015}. We set $T_{\min}$ and $T_{\max}$ to 0.1 and 10.0, respectively. 
All reported results are averaged over three independent runs with random seeds 1, 2, and 3. All experiments were performed on an
NVIDIA RTX A6000 ada with 48GB of memory.

\begin{table*}[t!]
\centering
\large
\caption{Comparison of accuracy and calibration performance using the CLIP-ViT/B16 backbone. Values are mean of 3 runs (std: acc 0.11, ece 0.15). The best and second-best results are highlighted in \textbf{bold} and \underline{underline}, respectively.}
\setlength{\tabcolsep}{5pt}
\renewcommand{\arraystretch}{1.15}
\centering
\resizebox{0.95\textwidth}{!}{%
\begin{tabular}{l|c|cccccccccc|c}
\toprule
\textbf{Method} & \textbf{Metric} 
& \textbf{Air} & \textbf{Calt} & \textbf{Car} & \textbf{DTD} & \textbf{SAT} 
& \textbf{FLW} & \textbf{Food} & \textbf{Pets} & \textbf{SUN} & \textbf{UCF} & \textbf{Avg.} \\
\midrule

\multirow{2}{*}{CLIP-ViT-B/16~\citep{clip}} 
& Acc. & 23.9 & 92.9 & 65.3 & 44.3 & 41.3 & 67.3 & 83.6 & 88.0 & 62.5 & 65.0 & 63.41 \\
& ECE  & 5.11 & 5.50 & 4.25 & 8.50 & 7.40 & 3.00 & 2.39 & 4.37 & 2.53 & 3.59 & 4.67 \\

\midrule
\multirow{2}{*}{TPT~\citep{tpt}} 
& Acc. & 23.4 & 93.8 & 66.3 & 46.7 & 42.4 & 69.0 & 84.7 & 87.1 & 65.5 & 67.3 & 64.62 \\
& ECE  & 16.8 & 4.51 & 5.16 & 21.2 & 21.5 & 13.5 & 3.98 & 5.77 & 11.3 & 13.0 & 11.67 \\

\midrule
\multirow{2}{*}{C-TPT~\citep{c-tpt}} 
& Acc. & 24.0 & 93.6 & 65.8 & 46.0 & 43.2 & 69.8 & 83.7 & 88.2 & 64.8 & 65.7 & 64.48 \\
& ECE  & 4.36 & 4.24 & 1.59 & 11.9 & 13.2 & 5.04 & 3.43 & 1.90 & 5.04 & 2.54 & 5.32 \\

\midrule
\multirow{2}{*}{O-TPT~\citep{o-tpt}}  
& Acc. & 23.64 & 93.95 & 64.53 & 45.68 & 42.84 & 70.07 & 84.13 & 87.95 & 64.23 & 64.16 & 64.12 \\
& ECE  & 3.68 & 3.80 & 1.78 & 7.88 & 12.98 & 3.87 & 1.46 & 1.90 & 4.93 & 2.34 & 4.46 \\

% \midrule
% \multirow{2}{*}{FPP~\citep{fpp}} 
% & Acc. & 24.75 & 93.25 & 66.65 & 46.14 & 50.66 & 69.43 & 84.31 & 87.30 & 64.08 & 67.08 & \underline{65.37} \\
% & ECE  & 7.26 & 5.85 & 2.00 & 7.52 & 5.19 & 2.67 & 1.92 & 2.63 & 3.22 & 3.04 & \underline{4.13} \\

\midrule
\cellcolor{gray!20}     
& \cellcolor{gray!20}Acc.
& \cellcolor{gray!20}23.64
& \cellcolor{gray!20}94.56
& \cellcolor{gray!20}67.44
& \cellcolor{gray!20}47.10
& \cellcolor{gray!20}43.09
& \cellcolor{gray!20}70.65
& \cellcolor{gray!20}85.17
& \cellcolor{gray!20}89.96
& \cellcolor{gray!20}66.94
& \cellcolor{gray!20}68.51
& \cellcolor{gray!20}\textbf{65.71} \\
\multirow{-2}{*}{\cellcolor{gray!20}Ours}    
& \cellcolor{gray!20}ECE
& \cellcolor{gray!20}3.94
& \cellcolor{gray!20}1.60
& \cellcolor{gray!20}2.67
& \cellcolor{gray!20}5.68
& \cellcolor{gray!20}9.56
& \cellcolor{gray!20}2.17
& \cellcolor{gray!20}1.76
& \cellcolor{gray!20}2.09
& \cellcolor{gray!20}3.61
& \cellcolor{gray!20}3.46
& \cellcolor{gray!20}\textbf{3.65} \\

\bottomrule
\end{tabular}
}
\label{tab:tpt_FG}
\vspace{-5pt}
\end{table*}

\begin{table*}[t!]
\centering
\caption{Comparison of accuracy and calibration error using the CLIP-ViT/B16 backbone within the TPT framework under predefined prompts trained from CoOp and MaPLe. }
\resizebox{0.95\textwidth}{!}{%
\begin{tabular}{l|c|cccccccccc|c}
\toprule
\textbf{Method} & \textbf{Metric} & \textbf{Air} & \textbf{Calt} & \textbf{Car} & \textbf{DTD} & \textbf{SAT} & \textbf{FLW} & \textbf{Food} & \textbf{Pets} & \textbf{SUN} & \textbf{UCF} & \textbf{Avg.} \\
\midrule
\multirow{2}{*}{CoOp+TPT}        & Acc. &  20.0 & 94.0 & 65.6 & 44.5 & 40.6 & 68.7 & 83.8 & 89.1 & 65.6 & 67.2 & \underline{63.91} \\
                                 & ECE & 29.6 & 3.65 & 6.63 & 34.8 & 31.3 & 19.9 & 9.66 & 7.40 & 20.8 & 19.9 & 18.36 \\
\midrule
\multirow{2}{*}{CoOp+C-TPT}     & Acc. &  19.2 & 93.9 & 63.1 & 45.0 & 40.7 & 69.0 & 83.7 & 89.3 & 65.1 & 66.6 & 63.56 \\
                                & ECE & 21.5 & 1.66 & 2.45 & 21.0 & 13.2 & 10.2 & 4.49 & 2.12 & 11.8 & 12.0 & 10.04 \\
\midrule
\multirow{2}{*}{CoOp+O-TPT}     & Acc. & 18.69 & 93.71 & 64.12 & 45.45 & 40.17 & 68.57 & 83.55 & 89.07 & 64.01 & 65.64 & 63.29 \\
                                & ECE & 16.82 & 0.92 & 2.85 & 16.02 & 13.76 & 6.81 & 3.59 & 1.92 & 7.23 & 9.16 & \underline{7.91} \\
% \midrule
% & Acc. & 21.51 & 93.96 & 63.74 & 44.80 & 48.72 & 68.57 & 84.31 & 88.42 & 66.44 & 67.30 & \textbf{64.78} \\
% \multirow{-2}{*}{CoOp+FPP }                             
% & ECE  & 9.34 & 3.49 & 5.33 & 11.79 & 9.69 & 6.00 & 1.21 & 2.00 & 3.00 & 4.88 & \underline{5.67}\\
\midrule
\cellcolor{gray!20}      
        & \cellcolor{gray!20}Acc. 
        & \cellcolor{gray!20}18.69 
        & \cellcolor{gray!20}93.79 
        & \cellcolor{gray!20}63.47 
        & \cellcolor{gray!20}44.15 
        & \cellcolor{gray!20}44.11 
        & \cellcolor{gray!20}70.04 
        & \cellcolor{gray!20}83.97 
        & \cellcolor{gray!20}89.51 
        & \cellcolor{gray!20}64.91 
        & \cellcolor{gray!20}67.25 
        & \cellcolor{gray!20}\textbf{63.99} \\
\multirow{-2}{*}{\cellcolor{gray!20}CoOp+Ours}                             
        & \cellcolor{gray!20}ECE  
        & \cellcolor{gray!20}4.19 
        & \cellcolor{gray!20}2.08 
        & \cellcolor{gray!20}7.66 
        & \cellcolor{gray!20}9.12 
        & \cellcolor{gray!20}9.08 
        & \cellcolor{gray!20}3.36 
        & \cellcolor{gray!20}1.77 
        & \cellcolor{gray!20}2.40 
        & \cellcolor{gray!20}3.33 
        & \cellcolor{gray!20}3.33 
        & \cellcolor{gray!20}\textbf{4.63} \\
\midrule
\midrule
\multirow{2}{*}{MaPLe+TPT}        & Acc. &  24.36 & 94.42 & 66.50 & 50.05 & 47.32 & 70.72 & 85.01 & 87.78 & 64.87 & 66.48 & \underline{65.75} \\
                                 & ECE & 10.58 & 2.38 & 4.14 & 11.80 & 9.42 & 11.63 & 1.78 & 1.79 & 8.47 & 7.41 & 6.94 \\
\midrule
\multirow{2}{*}{MaPLe+O-TPT}     & Acc. & 24.00 & 92.29 & 65.38 & 49.11 & 44.58 & 71.53 & 84.35 & 89.97 & 63.49 & 65.82 & 65.05 \\
                                & ECE & 6.41 & 3.49 & 3.61 & 4.90 & 7.92 & 4.35 & 1.49 & 3.97 & 2.78 & 2.22 & \underline{4.11} \\
% \midrule
       
%         & Acc. & 24.06 & 93.47 & 66.27 & 48.94 & 52.19 & 69.35 & 85.17 & 91.55 & 68.29 & 70.84 & \textbf{67.01} \\
% \multirow{-2}{*}{MaPLe+FPP }                            
%         & ECE  & 3.98 & 2.83 & 2.18 & 11.58 & 1.91 & 4.03 & 3.53 & 5.03 & 1.74 & 3.93 & \underline{4.07} \\
\midrule
\cellcolor{gray!20}      
        & \cellcolor{gray!20}Acc. &  \cellcolor{gray!20}22.55 & \cellcolor{gray!20}93.45 & \cellcolor{gray!20}66.11 & \cellcolor{gray!20}48.45 & \cellcolor{gray!20}48.34 & \cellcolor{gray!20}72.24 & \cellcolor{gray!20}85.00 & \cellcolor{gray!20}90.71 & \cellcolor{gray!20}65.68 & \cellcolor{gray!20}67.93 & \cellcolor{gray!20}\textbf{66.05} \\
\multirow{-2}{*}{\cellcolor{gray!20}MaPLe+Ours}                            
        & \cellcolor{gray!20}ECE  & \cellcolor{gray!20}2.12 & \cellcolor{gray!20}2.90 & \cellcolor{gray!20}4.55 & \cellcolor{gray!20}5.58 & \cellcolor{gray!20}5.32 & \cellcolor{gray!20}2.03 & \cellcolor{gray!20}3.53 & \cellcolor{gray!20}1.25 & \cellcolor{gray!20}1.55 & \cellcolor{gray!20}3.93 & \cellcolor{gray!20}\textbf{3.28} \\
\bottomrule
\end{tabular}
} % end resizebox
\label{tab:tpt_init}
% \vspace{-4.0pt}
\end{table*}

\subsection{Main Results}
\noindent\textbf{Fine-Grained Classification.}
We compare our method with existing test-time prompt tuning approaches on ten benchmark datasets. As shown in Table~\ref{tab:tpt_FG}, our method consistently improves both accuracy and calibration compared to prior methods.
In terms of accuracy, our method achieves an average accuracy of 65.71\%, outperforming TPT (64.62\%), C-TPT (64.48\%), and O-TPT (64.12\%). These results demonstrate that the proposed calibration-aware objective not only improves prediction reliability but also maintains competitive recognition performance under distribution shifts.
More importantly, our method significantly improves calibration. While TPT suffers from severe miscalibration with an average ECE of 11.67, our method reduces it to 3.65, achieving the best calibration among all compared methods. Compared to prior calibration-oriented approaches, our method achieves gains of 1.67 and 0.81 ECE, improving over C-TPT (5.32) and O-TPT (4.46). These results indicate that explicitly modeling uncertainty and controlling prediction confidence lead to substantially more reliable calibration.

% \vspace{3pt}
\noindent\textbf{Pre-trained Prompts.}
Table~\ref{tab:tpt_init} presents the performance of our method compared to existing TPT-based approaches under both CoOp and MaPLe initializations. 
Across all settings, our method consistently achieves the lowest ECE while maintaining competitive accuracy. 
Under the CoOp initialization, our method reduces the average ECE to 4.63, substantially outperforming prior approaches such as O-TPT (7.91). At the same time, it achieves the highest average accuracy of 63.99\%, demonstrating that improved calibration does not come at the expense of recognition performance.  
Notably, our method achieves substantial improvements in calibration across most datasets, demonstrating its effectiveness in mitigating overconfidence.
Similarly, under the MaPLe initialization, our method achieves the best calibration performance with an average ECE of 3.28, significantly improving over the previous best result of 4.11. Moreover, it attains the highest average accuracy of 66.05\% among all compared methods, showing that the proposed calibration-aware objective consistently improves both reliability and accuracy across different prompt initialization strategies.

% \vspace{3pt}
\noindent\textbf{Different TPT Framework.}
We further evaluate our method within the DynaPrompt~\citep{dynaprompt} framework to verify its generality under a most recent test-time prompt tuning strategy. 
As shown in Table~\ref{tab:dynaprompt_FG}, our method achieves the best calibration performance while maintaining strong accuracy.
Compared to existing approaches, our method significantly reduces the average ECE to 4.04, outperforming prior methods. 
At the same time, it achieves the highest accuracy (65.51\%), outperforming the original DynaPrompt framework, C-TPT, and O-TPT by 1.29\%, 1.75\%, and 2.16\%, respectively.
Notably, the calibration improvement is consistent across most datasets, indicating that our method effectively mitigates overconfidence.
These results demonstrate that our approach is not limited to a specific prompt tuning approach but can be integrated into different frameworks.

\begin{table*}[t!]
\centering
\caption{Comparison of accuracy and calibration error using the CLIP-ViT/B16 backbone within the DynaPrompt framework under a predefined hard prompt (``a photo of a'').}
\resizebox{0.95\textwidth}{!}{%
\begin{tabular}{l|c|ccccccccc|c}
\toprule
\textbf{Method} & \textbf{Metric} & \textbf{Air} & \textbf{Calt} & \textbf{Car} & \textbf{DTD} & \textbf{SAT} & \textbf{FLW} & \textbf{Food} & \textbf{Pets} & \textbf{UCF} & \textbf{Avg.} \\
\midrule

\multirow{2}{*}{DynaPrompt (Replication)}        
& Acc. & 22.68 & 94.16 & 66.88 & 47.87 & 35.91 & 69.67 & 84.92 & 87.71 & 68.17 & \underline{64.22} \\
& ECE & 18.67 & 3.05 & 5.08 & 23.19 & 33.76 & 12.73 & 6.75 & 5.62 & 13.72 & 13.62 \\
                                 
\midrule
\multirow{2}{*}{DynaPrompt+C-TPT}     
& Acc. & 23.34 & 93.83 & 66.04 & 46.99 & 36.57 & 70.08 & 83.58 & 88.61 & 65.74 & 63.86 \\
& ECE & 10.21 & 2.51 & 2.17 & 15.87 & 16.94 & 4.35 & 2.21 & 2.43 & 5.33 & 6.89 \\

\midrule
\multirow{2}{*}{DynaPrompt+O-TPT}     
& Acc. & 22.41 & 93.67 & 65.58 & 45.80 & 36.32 & 68.74 & 83.39 & 88.55 & 65.69 & 63.35 \\
& ECE & 9.60 & 2.74 & 2.13 & 14.31 & 17.73 & 3.60 & 2.03 & 2.78 & 3.43 &\underline{6.48} \\

% \midrule
% \multirow{2}{*}{DynaPrompt+FPP }
% & Acc. & 24.96 & 92.33 & 66.26 & 46.10 & 49.09 & 69.63 & 84.51 & 88.36 & 66.09 & \underline{65.26} \\
% & ECE  & 5.28 & 5.72 & 3.70 & 5.27 & 8.49 & 2.44 & 2.94 & 4.96 & 2.20 & \underline{4.56} \\

\midrule
\cellcolor{gray!20}      
        & \cellcolor{gray!20}Acc. & \cellcolor{gray!20}22.59 & \cellcolor{gray!20}93.43 & \cellcolor{gray!20}67.16 & \cellcolor{gray!20}46.99 & \cellcolor{gray!20}45.32 & \cellcolor{gray!20}71.12 & \cellcolor{gray!20}85.31 & \cellcolor{gray!20}89.55 & \cellcolor{gray!20}68.09 & \cellcolor{gray!20}\textbf{65.51} \\
\multirow{-2}{*}{\cellcolor{gray!20}DynaPrompt+Ours}                            
        & \cellcolor{gray!20}ECE  & \cellcolor{gray!20}4.12 & \cellcolor{gray!20}2.75 & \cellcolor{gray!20}4.81 & \cellcolor{gray!20}5.71 & \cellcolor{gray!20}9.10 & \cellcolor{gray!20}2.22 & \cellcolor{gray!20}2.03 & \cellcolor{gray!20}3.40 & \cellcolor{gray!20}2.19 & \cellcolor{gray!20}\textbf{4.04} \\
\bottomrule
\end{tabular}
}
\label{tab:dynaprompt_FG}
\vspace{-5pt}
\end{table*}
% \vspace{3pt}
\begin{table}[t!]
\centering
\caption{Comparison of accuracy and calibration error in natural distribution shift datasets.}
\vspace{8pt}
\resizebox{0.94\columnwidth}{!}{%
\begin{tabular}{l|c|ccccc|c}
\toprule
\textbf{Method} & \textbf{Metric} & \textbf{I} & \textbf{I-A} & \textbf{I-V2} & \textbf{I-R} & \textbf{I-S} & \textbf{OOD Avg.} \\
\midrule
\multirow{2}{*}{CLIP~\cite{clip}} 
& Acc. & 66.7 & 47.8 & 60.8 & 74.0 & 46.1 & 57.18 \\
& ECE & 2.12 & 8.61 & 3.01 & 3.58 & 4.95 & 5.04 \\
\midrule    
\multirow{2}{*}{TPT~\cite{tpt}} 
& Acc. & 69.0 & 52.6 & 63.0 & 76.7 & 47.5 & \textbf{59.95} \\
& ECE & 10.6 & 16.4 & 11.1 & 4.36 & 16.1 & 11.99 \\
\midrule         
\multirow{2}{*}{C-TPT~\cite{c-tpt}} 
& Acc. & 68.5 & 51.6 & 62.7 & 76.0 & 47.9 & 59.55 \\
& ECE & 3.15 & 8.16 & 6.23 & 1.54 & 7.35 & 5.82 \\
\midrule   
\multirow{2}{*}{O-TPT~\cite{o-tpt}}  
& Acc. & 67.3 & 49.9 & 61.7 & 72.6 & 47.1 & 57.82 \\
& ECE & 1.97 & 7.22 & 3.97 & 1.46 & 6.87 & \underline{4.88} \\
% \midrule   
% \multirow{2}{*}{FPP~\cite{fpp}}
% & Acc. & 67.8 & 52.3 & 61.9 & 76.7 & 47.2 & 59.53 \\
% & ECE & 2.97 & 5.38 & 3.45 & 7.07 & 2.62 & \underline{4.63} \\
\midrule   
\cellcolor{gray!20}
& \cellcolor{gray!20}Acc. 
& \cellcolor{gray!20}68.19 
& \cellcolor{gray!20}51.88 
& \cellcolor{gray!20}63.00 
& \cellcolor{gray!20}77.50 
& \cellcolor{gray!20}46.18 
& \cellcolor{gray!20}\underline{59.64} \\

\multirow{-2}{*}{\cellcolor{gray!20}Ours}
& \cellcolor{gray!20}ECE 
& \cellcolor{gray!20}3.42 
& \cellcolor{gray!20}5.37 
& \cellcolor{gray!20}4.38 
& \cellcolor{gray!20}3.30 
& \cellcolor{gray!20}1.30 
& \cellcolor{gray!20}\textbf{3.55} \\
\bottomrule
\end{tabular}
}
\label{tab:tpt_OOD}
\end{table}

\begin{table}[t]
  \centering
    \caption{Ablation study on the proposed temperature scaling and KLD averaged across fine-grained classification benchmarks.}
\vspace{8pt}
  \small
  \resizebox{0.8\columnwidth}{!}{%
    \begin{tabular}{%
      >{\centering\arraybackslash}p{0.25\linewidth}
      >{\centering\arraybackslash}p{0.25\linewidth}|
      >{\centering\arraybackslash}p{0.15\linewidth}
      >{\centering\arraybackslash}p{0.15\linewidth}
      >{\centering\arraybackslash}p{0.15\linewidth}}
      \toprule
      \textbf{Temperature Scaling} &
      \textbf{KLD} &
      \textbf{Accuracy} &
      \textbf{ECE}\\
      \midrule
      -- & -- & 64.52 & 9.80  \\
      \checkmark & -- & 64.77 & 8.70  \\
      -- & \checkmark & 64.97 & 6.18 \\
      \midrule
      \cellcolor{gray!20}\checkmark & \cellcolor{gray!20}\checkmark & 
      \cellcolor{gray!20}\textbf{65.71} & 
      \cellcolor{gray!20}\textbf{3.65}  \\
      \bottomrule
    \end{tabular}
  }
  \label{tab:ablation}
  \vspace{-5pt}
\end{table}

\begin{figure}[t]
\centering

\begin{minipage}[t]{0.59\columnwidth}
    \centering
    \includegraphics[width=\linewidth,
    trim=10pt 0pt 70pt 0pt,
    clip=true]{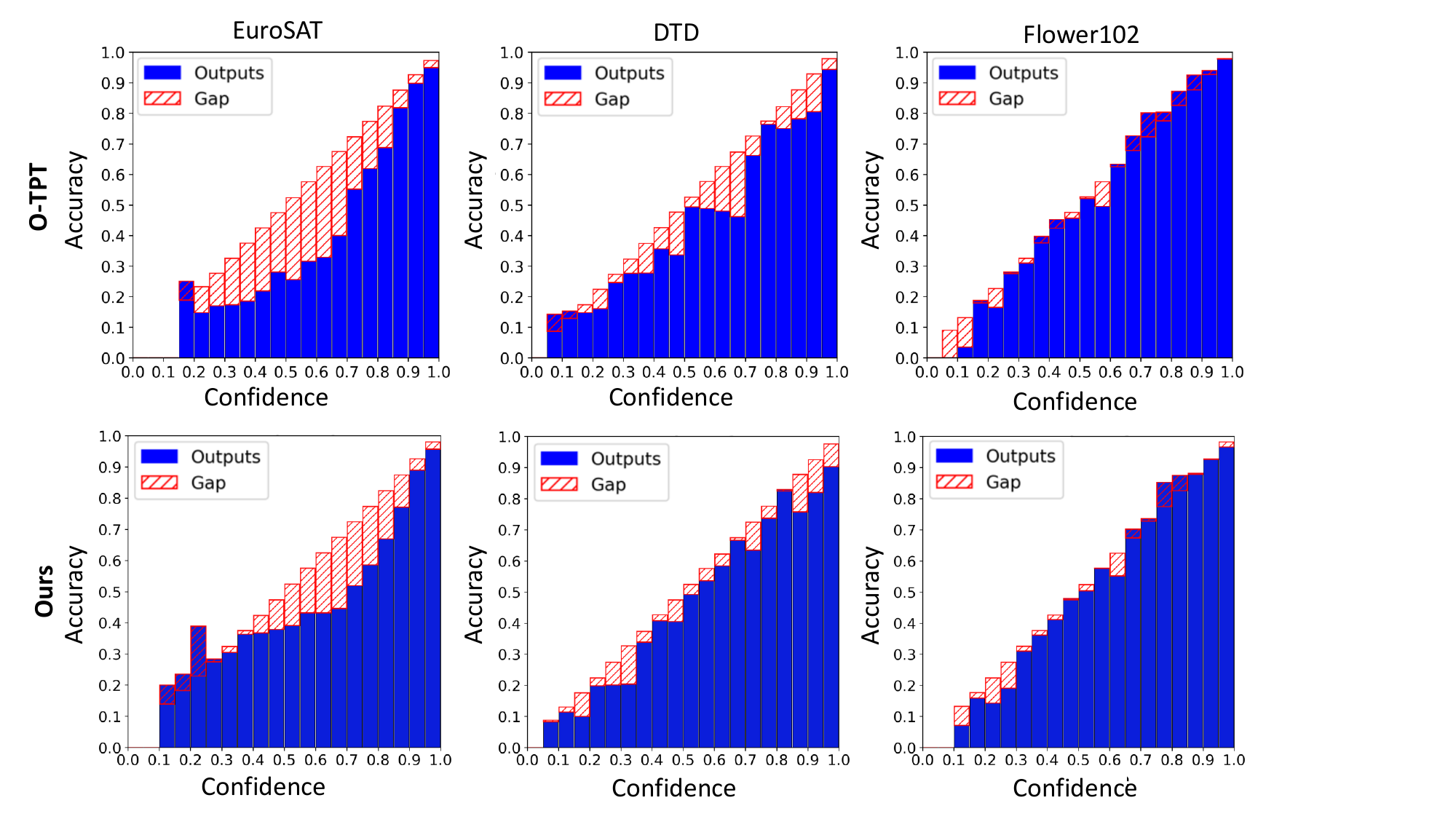}
    \caption{Reliability diagram showing improved calibration.}
    \label{fig:rd}
\end{minipage}
\hfill
\begin{minipage}[t]{0.4\columnwidth}
    \centering
    \includegraphics[height=0.9\linewidth,width=\linewidth,
    trim=0pt 0pt 0pt 0pt,
    clip=true]{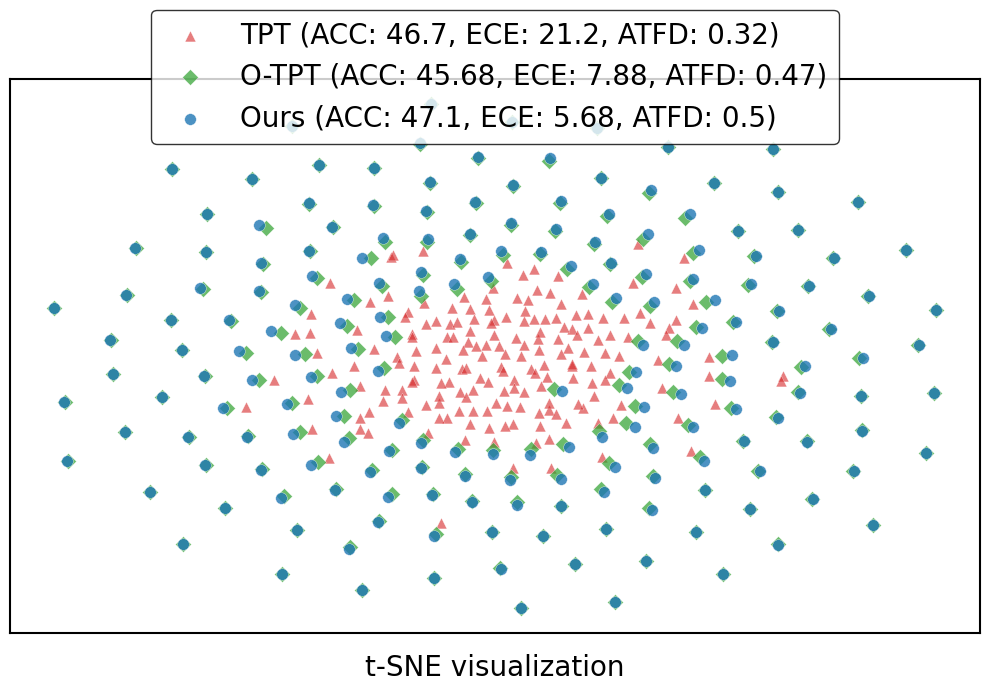}
    \caption{t-SNE visualization.}
    \label{fig:tsne}
    \vspace{-25pt}
\end{minipage}

\end{figure}

\noindent\textbf{Natural Distribution Shifts.}
% Tab.~\ref{tab:tpt_OOD} presents results on natural distribution shifts~\cite{Zhou2022b}, evaluating out-of-distribution (OOD) performance using widely adopted ImageNet variant datasets, where details provided in Appendix.
% In this setting, we fix $\lambda$ as 1.25 across all datasets.
% Under this configuration, our method consistently achieves notable improvements in both ECE and SCE, with only a minor accuracy reduction of 0.42\% compared to TPT.
% In contrast, O-TPT, which exhibits the best calibration among prior works, experiences an accuracy drop of more than 2\%.
% These findings highlight that our method maintains strong effectiveness under OOD scenarios.
We evaluate our method on ImageNet~\citep{imagenet} and its variants, including ImageNet-A~\citep{imagenet-a}, ImageNet-V2~\citep{imagenet-v2}, ImageNet-R~\citep{imagenet-r}, and ImageNet-Sketch~\citep{imagenet-sketch}, as shown in Table~\ref{tab:tpt_OOD}. 
While TPT achieves the highest accuracy, it suffers from severe miscalibration, resulting in a high average ECE of 11.99. 
In contrast, our method significantly improves calibration, achieving the lowest average ECE of 3.55, which substantially outperforms prior approaches. 
Importantly, this improvement is achieved without sacrificing accuracy, closely matching the best-performing baseline.

\subsection{More Studies}\noindent
\label{sec:morestudies}

\noindent\textbf{Ablation Study.}
We conduct an ablation study to analyze the contributions of our temperature scaling and KLD-based objective. As shown in Table~\ref{tab:ablation}, both components independently improve performance over the baseline, while their combination yields the best results.
Applying the temperature scaling alone provides a modest gain in accuracy (64.52\% → 64.77\%) and slightly improves calibration (ECE: 9.80 → 8.70), indicating that confidence smoothing partially alleviates overconfidence. In contrast, introducing the KLD loss leads to a more substantial improvement (Acc.: 64.97\%, ECE: 6.18), demonstrating that aligning predictions with augmentation-derived targets is more effective for calibration.
When combined, the two components achieve the best performance, reaching 65.71\% accuracy and reducing ECE to 3.65. This result shows that temperature scaling and the KLD objective are complementary: temperature scaling stabilizes prediction confidence, while the KLD loss provides a strong calibration signal. 

\noindent\textbf{Reliability Plots.}
To further analyze the calibration behavior, we present reliability diagrams on the EuroSAT, DTD, and Flower102 datasets in Fig.~\ref{fig:rd}. In a reliability diagram, perfect calibration is achieved when the prediction confidence matches the empirical accuracy within each confidence bin.
As shown in the figure, O-TPT exhibits noticeable calibration gaps across all datasets. In particular, both the EuroSAT and DTD datasets exhibit a tendency toward over-confidence.
In contrast, our method produces more well-aligned reliability curves across all datasets.

\noindent\textbf{Text Feature Dispersion.}
To examine how each method affects the geometry of class text representations, we visualize the learned text features using t-SNE and measure their dispersion with the average text feature dispersion (ATFD). As shown in Fig.~\ref{fig:tsne} , the baseline TPT exhibits relatively low dispersion (ATFD: 0.32), which corresponds to poor calibration performance (ECE: 21.20). 
O-TPT improves the dispersion of text features (ATFD: 0.47), resulting in a reduction in calibration error (ECE: 7.88). This observation aligns with a prior finding that increasing inter-class separation in the text space can enhance ECE performance~\citep{c-tpt}. 
Our method achieves the highest dispersion (ATFD: 0.50) while simultaneously improving both calibration (ECE: 5.68) and accuracy (47.10). This result demonstrates that our approach naturally encourages the dispersion of class text features, without requiring additional regularization losses.

\begin{figure}[t]
    \centering
    % 왼쪽: 두 개의 subfigure를 포함하는 영역
    \begin{minipage}[t]{0.64\linewidth}
        \centering
        \begin{subfigure}[t]{0.48\linewidth}
            \centering
            \includegraphics[width=\linewidth,, height=5cm]{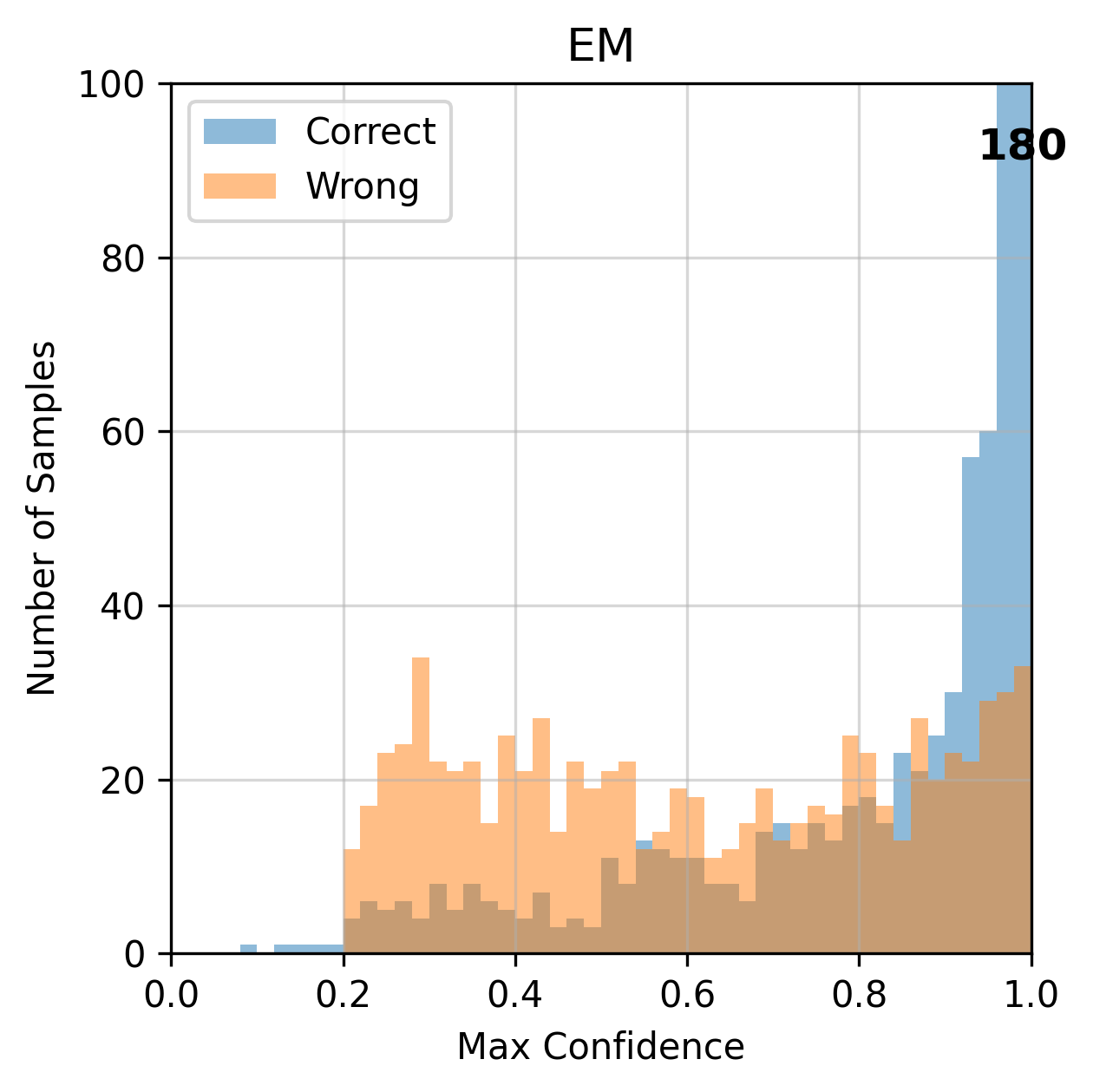}
            \caption{EM}
            \label{fig:maxconfem}
        \end{subfigure}
        \hfill
        \begin{subfigure}[t]{0.48\linewidth}
            \centering
            \includegraphics[width=\linewidth, height=5cm]{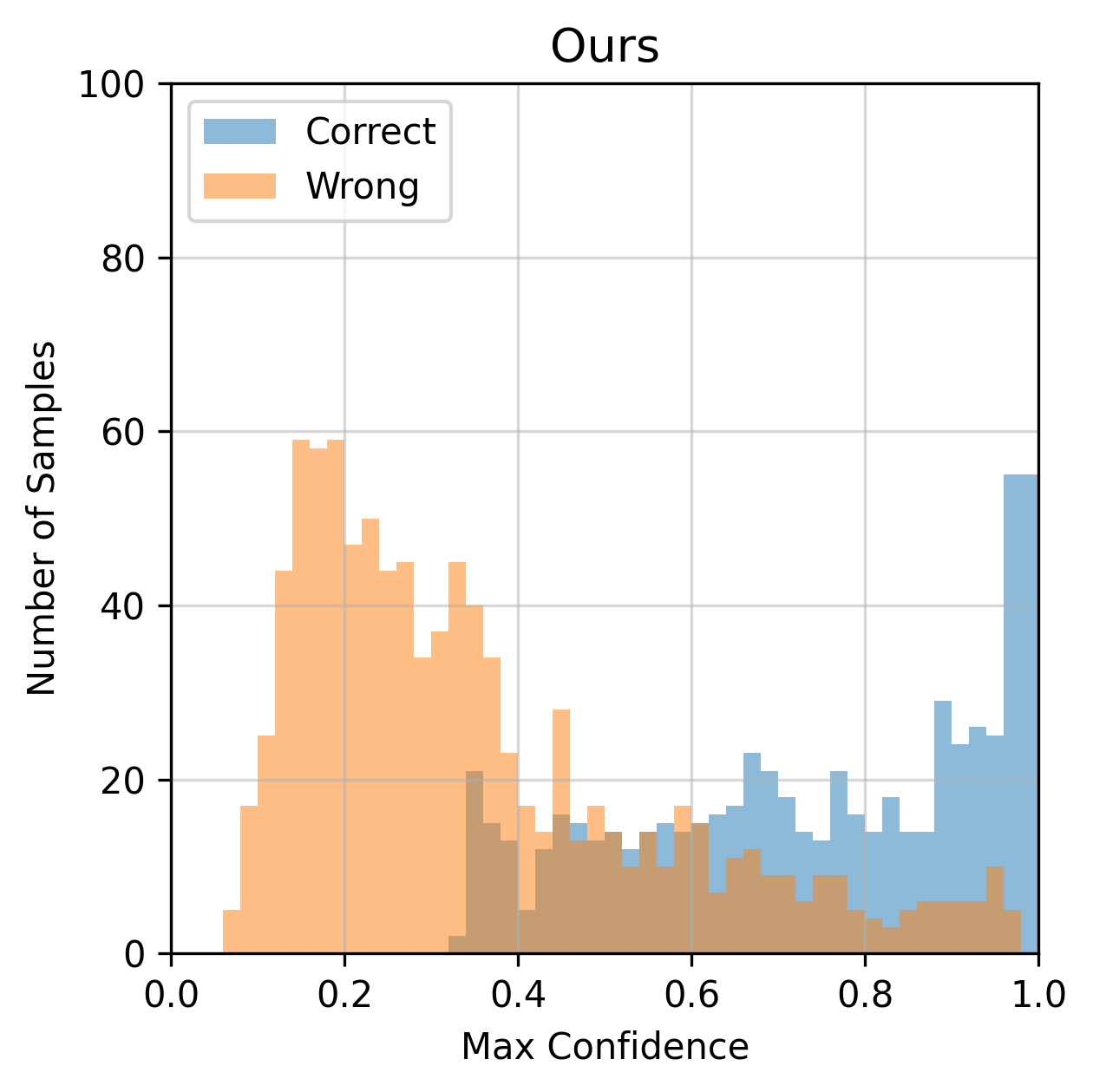}
            \caption{Ours}
            \label{fig:maxconfours}
        \end{subfigure}
        % \vspace{2pt} % 캡션과의 간격 조정
        \caption{Confidence distribution comparison between EM and our method. EM produces over-confident predictions regardless of correctness, while our method better separates correct and incorrect samples by reducing confidence for misclassified cases.}
        \label{fig:maxconf}
    \end{minipage}
    \hfill
    % 오른쪽: 단일 이미지를 포함하는 영역
    \begin{minipage}[t]{0.32\linewidth}
        \centering
        \vspace{-4.2cm}\includegraphics[width=\linewidth]{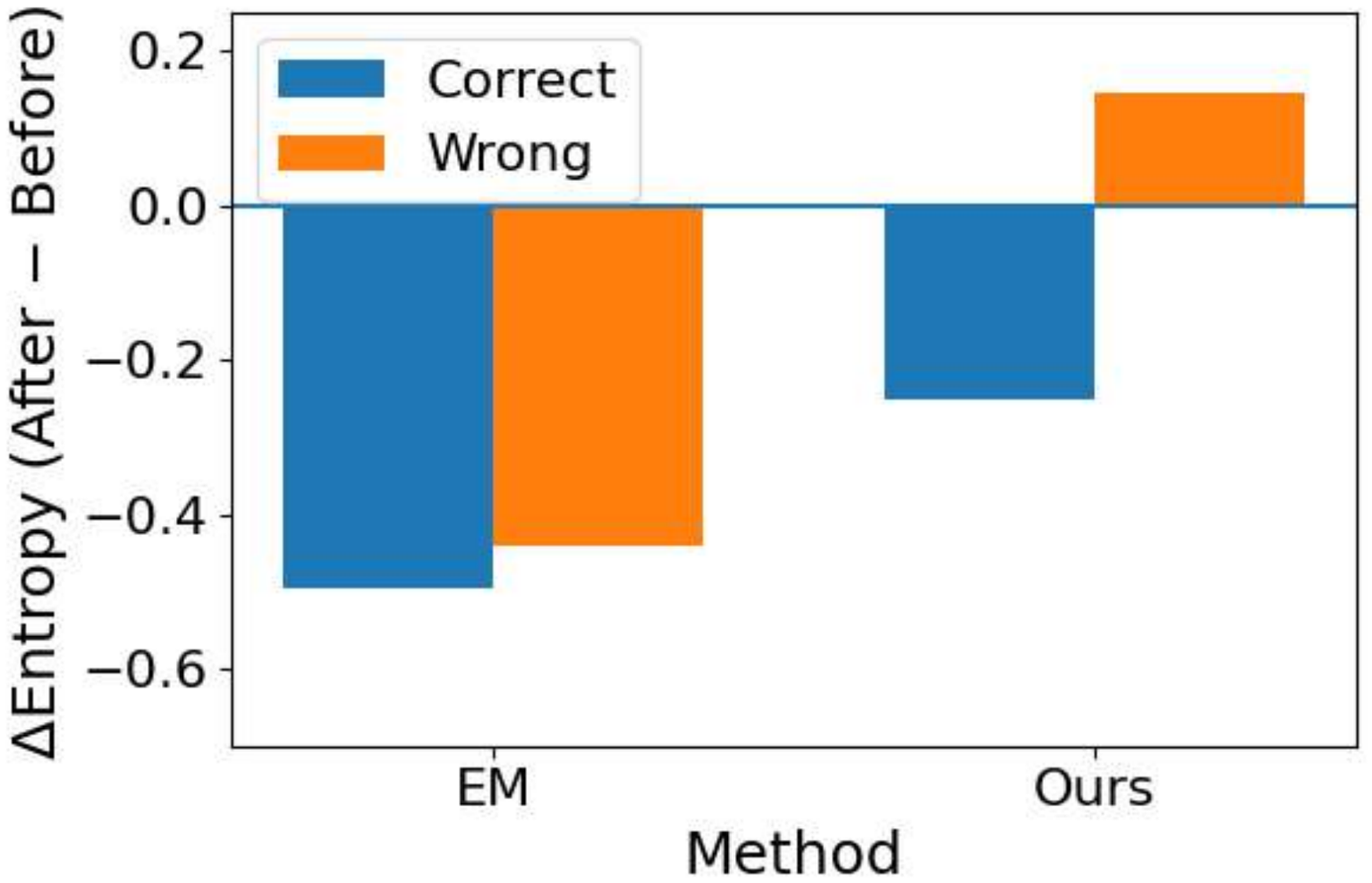}
        \vspace{-10pt} % 왼쪽 캡션 위치와 높이를 맞추기 위한 조정
        \caption{Entropy change before and after test-time adaptation. While EM reduces entropy for both correct and incorrect samples, our method increases entropy for incorrect samples while preserving confidence for correct ones.}
        \label{fig:entropy_delta}
    \end{minipage}
    \vspace{-10pt}
\end{figure}

\noindent\textbf{Confidence Distribution.}
We further analyze the distribution of prediction confidence to better understand calibration behavior. Figs.~\ref{fig:maxconfem} and \ref{fig:maxconfours} present the confidence distributions for correct and incorrect samples on the DTD dataset.
We observe that EM tends to collapse predictions toward high-confidence outputs regardless of correctness, resulting in over-confident predictions even for misclassified samples. In particular, the mean confidence of incorrect predictions remains relatively high ($\mu=0.609$), indicating that the model assigns excessive confidence to wrong predictions.
In contrast, our method exhibits a more balanced distribution, maintaining high confidence for correct predictions ($\mu=0.731$) while assigning significantly lower confidence to incorrect ones ($\mu=0.355$).
These results indicate that EM produces over-confident outputs irrespective of correctness, leading to poor calibration, whereas our method preserves uncertainty for difficult or ambiguous samples while maintaining reasonable confidence for correct predictions.

\paragraph{Entropy Behavior Analysis.}
To better understand the effect of our objective, we analyze the change in prediction entropy before and after test-time adaptation. Specifically, we measure $\Delta H = H_{\text{after}} - H_{\text{before}}$ for both correct and incorrect samples.
As shown in Fig.~\ref{fig:entropy_delta}, EM consistently reduces entropy for both correct and incorrect samples, with mean $\Delta H$ values of $-0.50$ and $-0.44$, respectively. This indicates that EM indiscriminately sharpens predictions, increasing confidence even when the prediction is incorrect.
In contrast, our method exhibits fundamentally different behavior. For correct samples, the entropy change is ($\Delta H \approx -0.25$), indicating that unnecessary over-sharpening is avoided. More importantly, for incorrect samples, our method increases entropy ($\Delta H \approx +0.14$), reflecting higher uncertainty in unreliable predictions.
These results demonstrate that our approach adaptively modulates prediction confidence: it preserves confidence for correct predictions while increasing uncertainty for incorrect ones.

\noindent
\begin{wrapfigure}{r}{0.3\linewidth}
    \vspace{-0pt}
    \centering
    \vspace{-0.3cm}\includegraphics[width=0.9\linewidth]{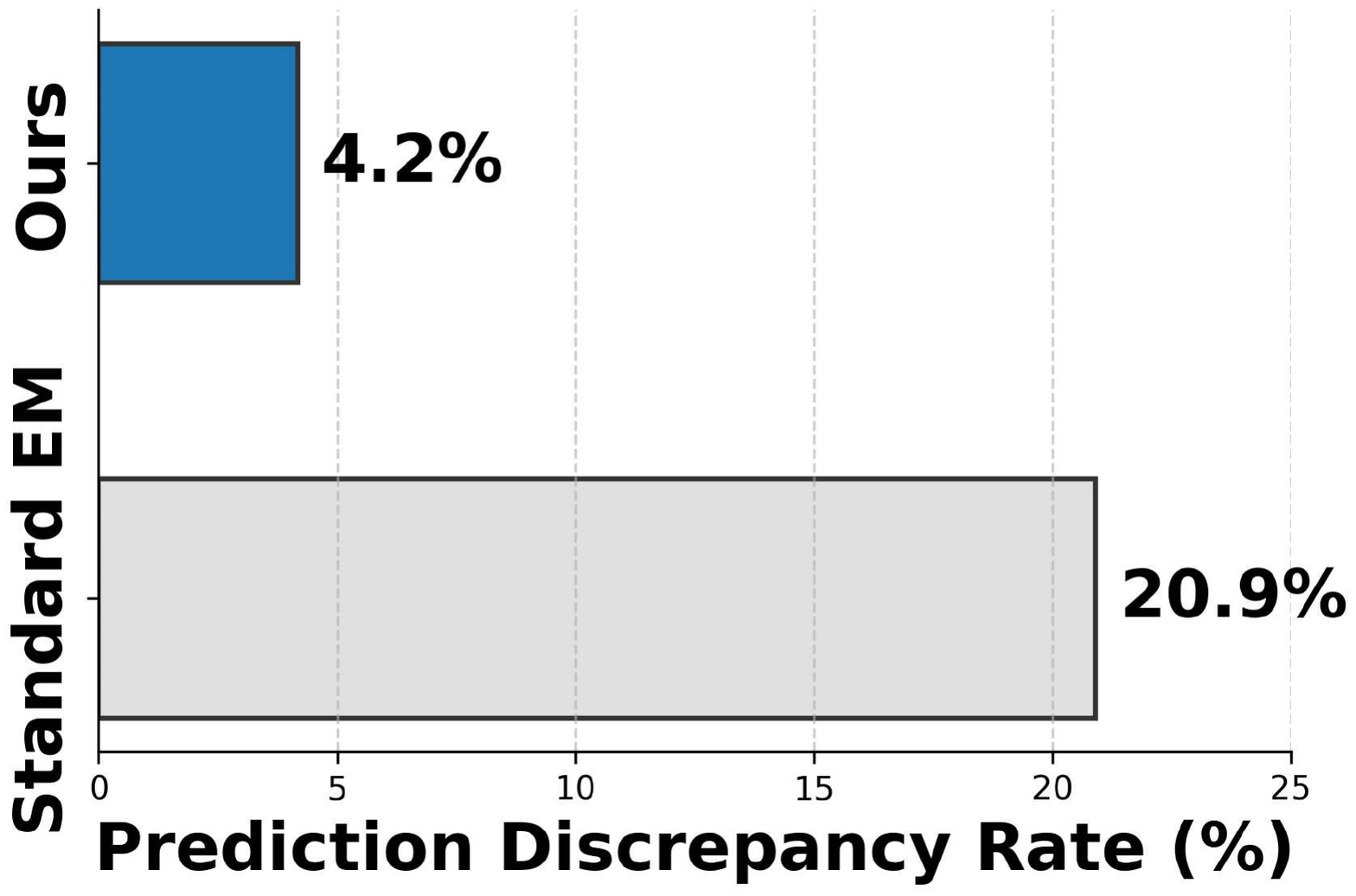}
    \caption{Prediction discrepancy between augmented and original views.}
    \label{fig:discrepancy}
    \vspace{-12pt}
\end{wrapfigure}
% \begin{wrapfigure}{r}{0.45\linewidth}
% \vspace{-10pt}
% \centering
% \includegraphics[width=\linewidth]{discrepancy_plot_horizontal.png}
% \end{wrapfigure}

\noindent\textbf{Analysis of Prediction Discrepancy.}
As shown in Fig.~\ref{fig:discrepancy} Standard EM exhibits a high prediction discrepancy rate of 20.9\%, where prediction discrepancy denotes the percentage of samples whose augmentation-derived prediction differs from the original-view prediction used for inference. This indicates that optimizing augmented-view probabilities can lead to significant misalignment with the actual inference-time prediction.
In contrast, our method substantially reduces this discrepancy to 4.2\%, demonstrating that directly aligning the original-view prediction with a confidence-aware target distribution enables more consistent and reliable control over prediction.

\section{Conclusion} 

In this paper, we identify a fundamental mismatch in entropy minimization based test-time prompt tuning.
%, where the optimization objective operates on augmented views while the final prediction is made from the original view, leading to suboptimal calibration. 
We further show that enforcing alignment without accounting for uncertainty can produce over-confident predictions, especially under ambiguous augmentation signals.
Building on these insights, we propose a calibration-aware objective that directly aligns the original-view prediction with a confidence-aware target distribution derived from augmented views. By incorporating adaptive temperature scaling, our method selectively enforces alignment when the target is reliable, while preserving uncertainty otherwise.
Notably, our approach improves calibration without requiring additional regularization.
% and can be seamlessly integrated into existing test-time prompt tuning frameworks. 
Extensive experiments demonstrate that our method consistently achieves better reliability while maintaining competitive accuracy across diverse benchmarks.  %ㅎㅇ

\paragraph{Limitations.}
Our experiments are primarily conducted on CLIP-based vision-language models, and evaluating the proposed method on a broader range of VLM architectures remains an important direction for future work. Additionally, the current adaptation process focuses on test-time inference, and exploring the scalability of our approach to extremely large-scale batches or real-time streaming data could provide further insights into its practical deployment.

\bibliographystyle{plainnat}
\bibliography{reference/main}
\newpage

\end{document}